\documentclass[11pt]{article}

\usepackage{acl}
\usepackage{cuted}

\usepackage{times}
\usepackage{latexsym}
\usepackage[T1]{fontenc}
\usepackage[utf8]{inputenc}
\usepackage{microtype}
\usepackage{inconsolata}

\usepackage{url}
\usepackage{booktabs}
\usepackage{amsmath}
\usepackage{amsfonts}
\usepackage{amssymb}
\usepackage{nicefrac}
\usepackage{graphicx}
\usepackage{subcaption}
\usepackage{xspace}
\usepackage{tabularx}
\usepackage{array}
\usepackage{multirow}

\newcolumntype{L}[1]{>{\raggedright\arraybackslash}p{#1}}
\newcolumntype{C}[1]{>{\centering\arraybackslash}p{#1}}
\newcolumntype{R}[1]{>{\raggedleft\arraybackslash}p{#1}}
\newcolumntype{Y}{>{\raggedright\arraybackslash}X}
\newcolumntype{Z}{>{\centering\arraybackslash}X}
\newcolumntype{W}{>{\raggedleft\arraybackslash}X}

\newcommand{\crafter}{\textsc{Crafter}\xspace}
\newcommand{\jericho}{\textsc{Jericho}\xspace}
\newcommand{\sciworld}{\textsc{ScienceWorld}\xspace}
\newcommand{\alfworld}{\textsc{ALFWorld}\xspace}

\usepackage[most]{tcolorbox}
\usepackage{xcolor}
\usepackage{cuted}

\definecolor{quanback}{HTML}{F7F8FB}
\definecolor{quanframe}{HTML}{6A7BA2}

\newtcolorbox{casebox}[1]{
  enhanced,
  breakable,
  colback=quanback,
  colframe=quanframe,
  boxrule=0.8pt,
  arc=2mm,
  left=2.5mm,
  right=2.5mm,
  top=1.6mm,
  bottom=1.6mm,
  title={#1},
  fonttitle=\bfseries,
  coltitle=black
}

\DeclareRobustCommand{\method}{\texorpdfstring{$\mathrm{S}^{3}$Mem}{S3Mem}}
\DeclareRobustCommand{\methodfull}{Structured Spatiotemporal Scene--Event Memory(\method{})}

\title{\texorpdfstring{S\textsuperscript{3}Mem}{S3Mem}: Structured Spatiotemporal
Scene-Event Memory for Long-Horizon Interactive Question Answering}

\author{
\begin{tabular}{c}
Encheng Su$^{1}$,
Jianyu Wu$^{2}$,
Jinouwen Zhang$^{3}$,
Qiucheng Yu$^{4}$,
Chen Tang$^{5}$,
Pengze Li$^{6}$ \\
Lintao Wang$^{7}$,
Aoran Wang$^{3}$,
Xinzhu Ma$^{8}$,
Shixiang Tang$^{3,\dagger}$,
Yizhou Wang$^{5,\dagger}$,
Houqiang Li$^{1}$
\end{tabular}
\\[0.6em]
\begin{tabular}{c}
$^{1}$University of Science and Technology of China \quad
$^{2}$Shanghai Jiao Tong University \\
$^{3}$Shanghai AI Laboratory \quad
$^{4}$City University of Hong Kong \quad
$^{5}$The Chinese University of Hong Kong \\
$^{6}$Fudan University \quad
$^{7}$The University of Sydney \quad
$^{8}$Beihang University
\end{tabular}
\\[0.4em]
$^{\dagger}$Corresponding authors
}

\begin{document}

\maketitle

\begin{strip}
  \centering
  \vspace{-0.5cm}
  \includegraphics[width=\textwidth]{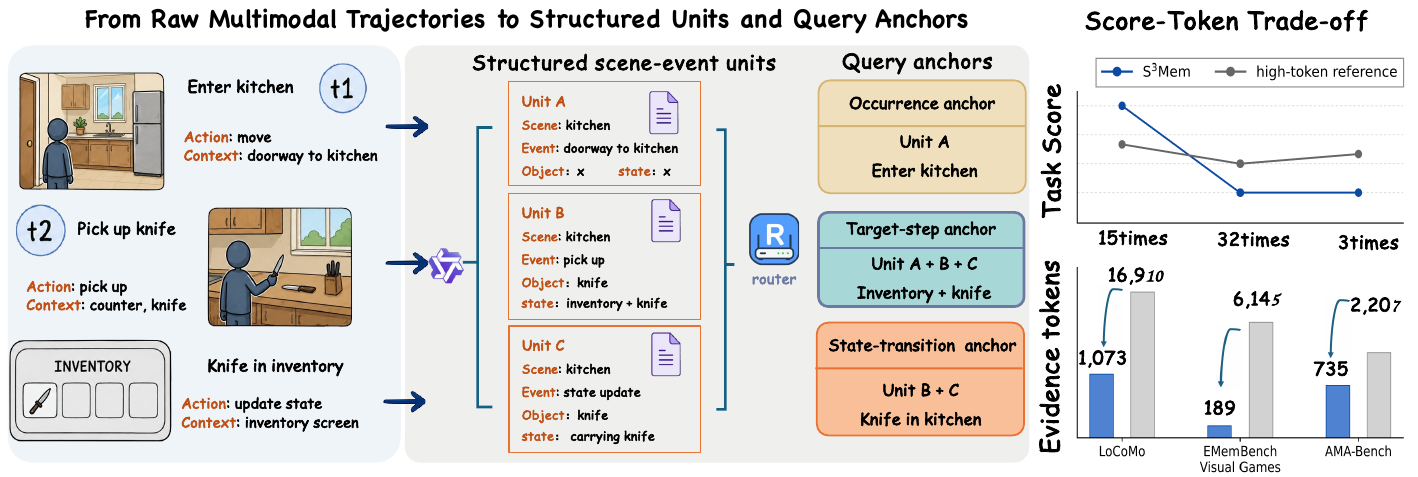}
  \captionof{figure}{
Overview of \method{} and its score--token trade-off.
\textbf{Left:} raw multimodal trajectories are written into structured scene--event units, and query anchors guide routing and evidence packing.
\textbf{Right:} across three memory benchmarks, \method{} achieves competitive task scores with substantially fewer reader-visible evidence tokens than high-token references.
  }
  \label{fig:method_overview}
\end{strip}

\begin{abstract}
Long-horizon memory question answering often requires sparse evidence from heterogeneous histories, including events, object states, visual observations, temporal relations, and causal steps. Existing memory interfaces expand reader context, retrieve semantically related chunks, or expose graph neighborhoods, but they are not explicitly designed to select compact evidence for a fixed reader. We propose Structured Spatiotemporal Scene--Event Memory (\method{}), a query-time memory interface that writes textual, visual, and agent-use histories into structured scene--event units and routes compact evidence packs to the reader. Its router scores candidate units, query anchors, and anchor--support links, enabling both single-hop selection and short multi-hop evidence chains without reader fine-tuning or test-time training. Across LoCoMo, EMemBench Visual Games, and AMA-Bench, \method{} provides a strong score--token trade-off, with the clearest gains on localized event, state, temporal, causal, or provenance evidence. On LoCoMo, \method{} reaches \(0.48\)  F1 and \(0.40\) BLEU with (1{,}073) evidence tokens per question, about \(15.8\times\) fewer than the LoCoMo reference. On EMemBench Visual Games, it obtains the best F1 and second-best accuracy with only  \(189\)tokens.On AMA-Bench, it is not the highest-scoring method, but remains competitive while using the fewest reader-visible evidence tokens.
\end{abstract}

\section{Introduction}
\label{sec:intro}

LLM-based agents increasingly operate over long-running workflows, including
coding~\citep{jimenez2024swebench,yang2024sweagent}, tool
use~\citep{qin2024toolllm,liu2024agentbench}, visual
interaction~\citep{koh2024visualwebarena,xie2024osworld}, and personal
assistance~\citep{li2024personalllmagents,zhong2024memorybank,park2023generative,shinn2023reflexion,packer2023memgpt,wang2023voyager}.
In these settings, later questions often depend on sparse evidence from earlier
interactions, such as an event, an object state, a visual observation, a temporal
relation, a source record, or a causal step.
The challenge is therefore not only to store long histories~\citep{bai2024longbench,liu2024lost}, but also to decide
which parts of the history should be exposed to the answer-time
\emph{reader}\footnote{In open-domain QA, a \emph{reader} typically denotes the
model or module that produces the final answer from given evidence or
context~\citep{chen2017reading,karpukhin2020dense}. We use the term in this
sense.}.

Existing long-horizon memory methods address this problem through several
interfaces. \textbf{Long-context prompting} places more interaction history into
the model context~\citep{bai2024longbench,liu2024lost}, but shifts the search burden to the reader and spends tokens
on irrelevant records. \textbf{Retrieval-based memory} indexes histories as
discrete chunks and retrieves them by semantic similarity~\citep{lewis2020retrieval},
but semantic relevance does not guarantee answerability. \textbf{Graph-based
memory} organizes histories through entities and relations~\citep{he2024gretriever},
but retrieved neighborhoods may still be too coarse or too broad for the
question. \textbf{Compression-based memory} summarizes long histories into
compact representations, but may discard critical details such as timestamps,
state transitions, visual cues, or source information. As a result, the reader
may receive redundant context or incomplete evidence.
These limitations indicate that the bottleneck of long-horizon memory is not just storage, but reader exposure. The key difficulty is a granularity gap: later questions often require fine-grained evidence, whereas existing systems expose coarse chunks, graph neighborhoods, or summaries.


To bridge this gap, we propose the Structured Memory Writer, \emph{e.g.}, a frozen Qwen or Qwen-VL model~\citep{yang2024qwen2,bai2025qwen25vl}, to map raw observations, interactions, and records into spatiotemporal event-state units. Each unit records time, event or observation, objects/entities, location or source, state change, relations, provenance, and a raw-record pointer. This shared schema turns textual, visual, and agent-use histories into a structured evidence space, allowing later modules to select compact support rather than broad retrieved context.

To operate over these structured units, we introduce \method{}, a plug-and-play query-time evidence interface for a frozen LLM reader. Given a memory-dependent question, \method{} first extracts lightweight query anchors and constructs a high-recall candidate pool from complementary semantic, temporal, visual, causal, and provenance cues. It then applies a lightweight evidence router with support, anchor, and edge heads to score candidate units and unit pairs, separating answer-bearing evidence from records that are only topically related. For single-hop questions, the packer selects one or a small set of high-support units, such as an event, state, frame, or source record. For multi-hop questions, it starts from a high-confidence anchor and expands through high-scoring edges to form a short anchor--support chain. Finally, a deterministic packer and reader harness render only the selected evidence to the reader, without test-time fine-tuning or access to gold evidence, labels, benchmark scores, or reader outputs.

We extensively validate the effectiveness and generality of \method{} on a broad suite of long-horizon memory tasks spanning text, visual, and agent-use settings. Across all domains, \method{} consistently demonstrates a superior trade-off between answer quality and evidence compactness.
\textbf{For text memory,} LoCoMo evaluates long conversations, documents, and textual histories, where \method{} achieves an official score of around \(0.48\) while using only about \textbf{$\frac{1}{15}$} of the token budget required by heavier RAG methods (Figure~\ref{fig:method_overview} right).
\textbf{For visual memory,} on EMemBench,
which demands rigorous tracking of frame-level observations and object state changes,
\method{} reaches about \(0.36\)  score with roughly \(189\) evidence tokens on visual trajectory benchmarks, corresponding to only about 3\% of AMem's and 7\% of Mem0's evidence-token budget.
\textbf{For agent-use memory,} AMA-Bench tests action traces, objectives, causal steps, and task-state changes, where \method{} obtains about \(0.39\) with about \(735\) evidence tokens, approaching graph-path accuracy and surpassing recent memory baselines, \emph{i.e.}, LightMem\citep{lightmem2025}, A-MEM\citep{amem2025agentic}, and MemoryOS\citep{memoryos2025}.
As an additional qualitative finding, the learned router exhibits strong transfer: it is trained on our constructed weak-supervision data rather than on the downstream benchmarks, yet generalizes well to long-horizon tasks.
Together, these results demonstrate that \method{} generalizes across heterogeneous memory sources and question types by exposing compact structured evidence rather than merely compressing context.



Our contributions are three-fold: (1) We introduce a structured spatiotemporal event-state memory representation for long-horizon histories, mapping observations, interactions, and records into units with time, event, object/entity, location/source, state change, relation, source metadata, and raw-pointer fields. (2) We propose \method{}, a lightweight evidence-routing framework that selects compact evidence before the reader answers. It handles single-hop questions by selecting one answer-bearing unit and multi-hop questions by constructing an anchor-support evidence chain. (3) We validate \method{} across visual memory, text memory, and agent-use memory using EMemBench visual/external, LoCoMo, and AMA-Bench. 
Our evaluation shows that \method{} provides a strong score--token trade-off across the three settings, with the clearest gains on text and visual memory and a compact competitive operating point on agent-use memory.


\begin{figure*}[t]
  \centering
  \includegraphics[width=\textwidth]{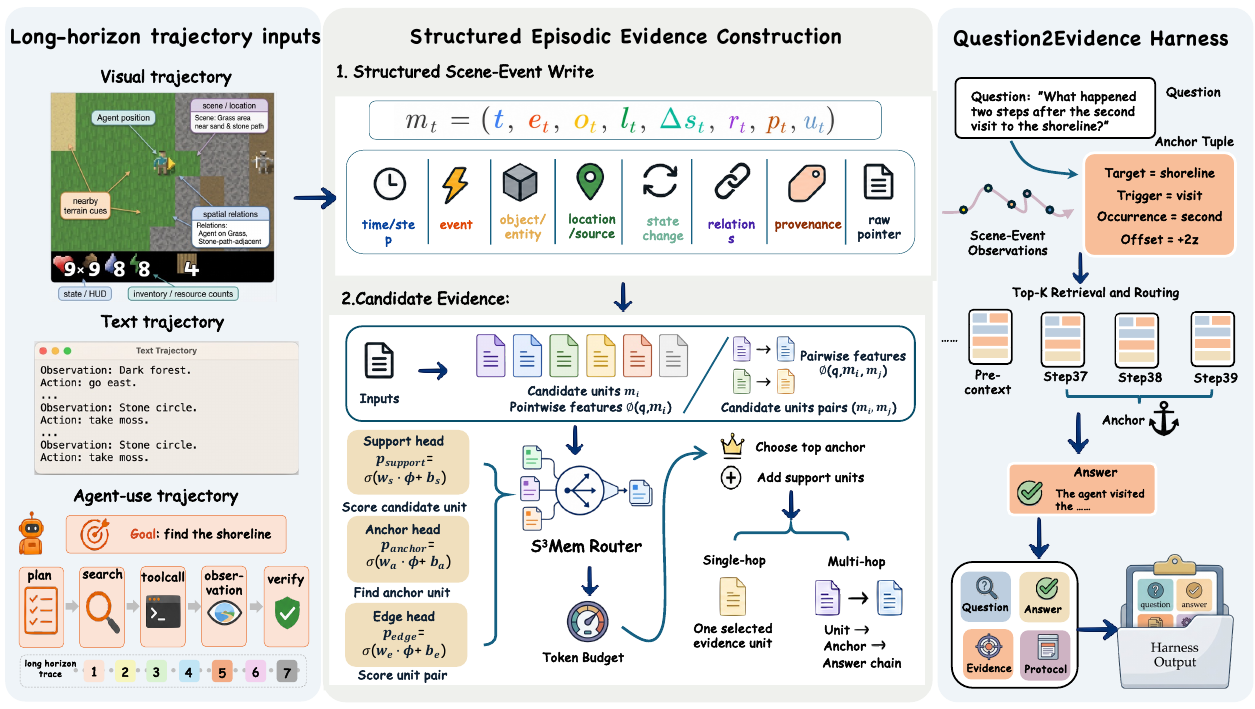}
    \caption{
    Overview of the \method{} pipeline.
    Left: visual, textual, and agent-use trajectories provide heterogeneous long-horizon memory sources.
    Middle: raw records are written into structured scene--event units, and candidate units or unit pairs are scored by support, anchor, and edge routing heads.
    Right: query anchors guide evidence selection, after which the packer constructs a compact single-hop unit or multi-hop anchor--support chain for the frozen answer-time reader.
    }
  \label{fig:overview}
\end{figure*}

\section{Related Work}
\label{sec:related}

\noindent\textbf{Retrieval-based memory and RAG.}
Sparse, dense, and retrieval-augmented generation methods reduce the amount of context shown to a language model by selecting external passages or memory chunks~\citep{lewis2020retrieval,guu2020realm,izacard2023atlas}.
Graph-based and relation-aware RAG further preserve entity, relation, and path information that flat retrieval often loses~\citep{he2024gretriever,arigraph2024}.
These methods are useful for long-horizon memory because they provide high-recall candidate evidence.
However, retrieved relevance is not the same as answerability.
A top-ranked chunk may mention the right object but the wrong occurrence, while a retrieved subgraph may include a related neighborhood without identifying the exact temporal offset, state transition, or support chain needed by the reader.
\method{} therefore treats retrieval and graph relations as candidate-proposal signals rather than final memory selection.
The router operates over structured event-state units and decides which compact single-hop unit or multi-hop anchor-support chain should actually be shown to the reader.

\noindent\textbf{Long-term agent memory.}
Recent agent-memory systems study how agents store, summarize, reflect on, update, and retrieve long-term memories~\citep{park2023generative,shinn2023reflexion,packer2023memgpt,wang2023voyager,amem2025agentic,memoryos2025,lightmem2025}.
These systems support persistent personalization, lifelong adaptation, task experience reuse, and efficient memory-augmented generation.
They are complementary to our work, but their main focus is often memory writing, consolidation, retrieval, or agent behavior over time.
\method{} focuses on a narrower query-time question:
when a user asks a memory-dependent question, which few memory units should be exposed, how many evidence tokens should be spent, and whether the selected units are sufficient for a fixed reader to answer.
We therefore compare against recent memory-system neighbors when they can be adapted to the same evidence-selection protocol, without claiming that such adaptations reproduce every deployment setting of the original systems.

\noindent\textbf{Structured episodic and event memory.}
Structured and episodic memory have long been studied in cognitive modeling and reinforcement learning~\citep{tulving1972episodic,nuxoll2007extending,blundell2016model,pritzel2017neural}, and recent LLM-agent memory systems revisit structured event memory~\citep{seem2026structured,arigraph2024}.
These works motivate representing memory with events, objects, locations, states, relations, and provenance rather than plain text alone.
Our difference is how this structure is used.
Prior structured-memory systems mainly use structure for richer storage or relation-preserving retrieval.
\method{} uses structure as a query-time routing substrate:
event fields identify occurrences, state fields expose before/after changes, temporal and causal links support multi-hop chains, visual fields connect frame-level evidence, and source fields support provenance-aware selection.
Thus, the structured memory unit is not only a record format, but the search space over which compact answerable evidence is selected.

Across these areas, prior work improves how information is retrieved, structured, stored, summarized, or orchestrated.
\method{} targets a different interface: the query-time decision of which compact structured evidence should be shown to a fixed reader.
This makes answerability and evidence cost first-class quantities rather than side effects of retrieval size, context length, or reader strength.

\section{Method}
\label{sec:method}


While existing memory systems can store extensive interaction histories, their ability to precisely isolate sparse, answer-bearing evidence for the reader remains limited. To unlock efficient long-horizon reasoning for any fixed LLM reader \(R\), we introduce \methodfull{}, a query-time evidence-routing framework (Fig.~\ref{fig:overview}). \method{} comprises a \textbf{structured memory writer} (Sec.~\ref{sec:structured_units}) that serializes heterogeneous raw histories into granular scene-event units, a lightweight \textbf{evidence router} (Sec.~\ref{sec:evidence_routing}) that dynamically scores these units, and an \textbf{evidence packer} with the reader harness (Sec.~\ref{sec:evidence_harness}) that seamlessly assembles them into single-hop facts or multi-hop anchor-support chains. Constructing compact, structurally cohesive evidence packs, \method{} is able to facilitate any fixed LLM reader for long-horizon interactive question answering.



\subsection{Structured Memory Writer}
\label{sec:structured_units}

Direct retrieval over unstructured raw histories often misses fine-grained state changes or causal steps. To bridge this gap, \method{} employs a frozen LLM writer, \emph{e.g.}, Qwen, as a Structured Memory Writer \(W\) to serialize the raw history \(H=(x_1,\ldots,x_T)\) into
structure memory units $M=(m_1,\cdots,m_T)$:
\begin{equation}
     M = W(H;\pi_{\mathrm{write}}),
\end{equation}
where $\pi_{\mathrm{write}}$ is the structured writing prompt given in Appendix~\ref{app:writer_prompt}.
For each raw segment \(x_t\), the writer extracts one or more structured scene--event memory units:
\begin{equation}
    m_i = (t_i, e_i, o_i, \ell_i, \Delta s_i, r_i, p_i, u_i).
\end{equation}

These fields explicitly capture the who-what-where-how of an occurrence: time (\(t_i\)), event/observation (\(e_i\)), objects/entities (\(o_i\)), location/source context (\(\ell_i\)), state changes (\(\Delta s_i\)), relations (\(r_i\)), provenance metadata (\(p_i\)), and a raw pointer (\(u_i\)) to the raw record.
Rather than simply compressing or overwriting the history, this design acts as a multi-dimensional semantic index, exposing highly searchable attributes while maintaining a direct pointer to the original evidence via (\(u_i\)).

Crucially, this event-state schema establishes a unified representation across fundamentally heterogeneous memory domains. While the underlying schema remains identical, it adaptively captures the most salient signals of each modality: textual histories emphasize entities and source continuity; visual streams highlight cross-frame object states and spatial changes; and agent-use traces focus on actions, tool traces, and causal updates. By projecting disparate data types into a shared representation space, the Structured Memory Writer enables the downstream evidence router to execute a single, domain-agnostic routing logic without requiring modality-specific architectures.


\noindent\textbf{Question Anchors.} To fully exploit this unified memory space, \method{} symmetrically processes the incoming question \(q\) by extracting lightweight query anchors. A query anchor structurally decomposes the intent of the question, identifying whether the reader seeks a specific event occurrence (\(e_i\)), a target state transition (\(\Delta s_i\)), or specific source provenance (\(p_i\)). Importantly, these anchors are parsed zero-shot directly from the question text, since they contain no gold labels and leak no ground-truth answers. They act strictly as directional routing cues. By projecting the unstructured query into the same granular schema as the memory units, these anchors constrain candidate construction and empower the downstream router to cleanly separate truly answer-bearing evidence from related noise.

\subsection{Evidence Router}
\label{sec:evidence_routing}


To circumvent the bottleneck of monolithic vector search, \method{} first assembles a high-recall candidate pool \(C(q,M)\) through multi-channel proposal. Rather than filtering early, this stage aggressively unites signals from semantic retrieval, anchor matching, visual continuity, and explicit causal links. The resolution of this noisy, heterogeneous pool is deliberately deferred to our Evidence Router.

Operating as a surgical filter, the Evidence Router then evaluates candidates directly through their structured attributes.
For each candidate unit \(m_i\), it computes pointwise features \(\phi(q,m_i)\), including retrieval score, anchor match, entity or object overlap, temporal position, source match, state-change indicators, and provenance fields.
For each candidate pair \((m_i,m_j)\), it computes pairwise features \(\psi(q,m_i,m_j)\), including temporal distance, shared entities, relation type, causal dependency, visual continuity, and source continuity. These features are processed by three highly efficient, lightweight scoring heads:
\begin{equation}
    p_{\mathrm{sup}}(m_i\mid q)=\sigma(w_s^\top \phi(q,m_i)+b_s),
\end{equation}
\begin{equation}
    p_{\mathrm{anc}}(m_i\mid q)=\sigma(w_a^\top \phi(q,m_i)+b_a),
\end{equation}
\begin{equation}
    p_{\mathrm{edge}}(m_i,m_j\mid q)=\sigma(w_e^\top \psi(q,m_i,m_j)+b_e).
\end{equation}

These three probabilities collectively define a dynamic routing topology.
The support probability ($p_{\mathrm{sup}}$) estimates whether a unit contains answer-relevant evidence.
The anchor probability ($p_{\mathrm{anc}}$) demonstrates whether a unit matches the query anchor, while
the edge probability ($p_{\mathrm{edge}}$) validates whether two units should be selected together as an anchor--support link in a multi-hop evidence chain.

\noindent\textbf{Discussion.} The core design philosophy of \method{} is that \textit{relatedness does not equal sufficiency}, so that we should disentangle semantic relevance from logical answerability. In standard RAG, a text chunk discussing a tool's general usage might exhibit high cosine similarity to a query, yet fail to capture the specific runtime state needed to answer it. Conversely, graph-based methods often blindly expand entire structural neighborhoods, drowning the reader in redundant context. By explicitly modeling anchor, support, and edge probabilities over structured fields, \method{} acts as a precise routing mechanism. It bypasses superficial semantic overlap to isolate only the minimal, logically coherent evidence chain required by the fixed reader.

\subsection{Evidence Packer and Reader Harness}
\label{sec:evidence_harness}

The continuous probabilities generated by the router cannot be directly consumed by a frozen LLM reader, which expects a concise and logically ordered context.
We therefore employ an Evidence Packer with the Reader Harness to convert router outputs into a structured
evidence pack.

\paragraph{Dynamic Evidence packing.}
Given the candidate pool \(C(q,M)\), support scores \(p_{\mathrm{sup}}\),
anchor scores \(p_{\mathrm{anc}}\), and edge scores \(p_{\mathrm{edge}}\), the
packer constructs
\begin{equation}
E =
\operatorname{Pack}
\left(
q,\,
C(q,M),\,
p_{\mathrm{sup}},
p_{\mathrm{anc}},
p_{\mathrm{edge}}
\right).
\end{equation}
The packer is fixed at inference time and has no learned parameters. 
For single-hop questions, the packer selects one or a small number of high-support units, such as an event, state, frame, source record, or textual fact.
For multi-hop questions, it first selects a high-confidence anchor and then expands to connected support units using edge scores. To maximize token efficiency, redundant units (e.g., overlapping source spans or duplicated state updates) are aggressively skipped. Finally, the selected units are logically ordered to match the query intent: temporal queries are sorted chronologically, and causal queries are arranged topologically.


\paragraph{Reader Harness.}
The selected pack \(E\) is rendered into a benchmark-compatible reader prompt:
\begin{equation}
    \hat{a} = R(\mathrm{Harness}(q,E)).
\end{equation}
The harness acts strictly as a lightweight translation layer. It serializes the structured fields of the selected units into a benchmark-compatible natural language prompt and appends the question $q$. It does not retrieve additional records, modify router scores, call another reasoning model, or access gold answers, gold evidence, judge labels, benchmark scores, or reader outputs. Thus, the evaluated component is the memory-side evidence interface: which structured units are selected and how they are organized before answering.

\subsection{Training and Inference Boundary}
The writer, router, and reader play separate roles. The writer is a frozen serializer that converts raw histories into structured units using the prompt in Appendix~\ref{app:writer_prompt}. The reader is also fixed and answers only from the rendered evidence pack. When supervision is available, only the lightweight router is trained offline from weak evidence labels, which supervise support-unit prediction, anchor prediction, and anchor--support edge prediction. During benchmark inference, all learned router parameters are fixed. The system selects evidence using only the question, written memory units, candidate features, and pre-answer relations without any test-time fine-tuning or benchmark-specific leakage.


\section{Experiments}
\label{sec:experiments}

\subsection{Experimental Setup}
\label{sec:experimental_setup}


We evaluate \method{} as a query-time memory interface for long-horizon question answering.
Given a question and a memory source, each method provides the context or evidence used by a fixed answer-time reader.
Following recent long-term memory evaluations~\citep{amem2025agentic,lightmem2025}, we evaluate both answer quality and the number of evidence tokens exposed to the reader.

\paragraph{Benchmarks.}
We evaluate \method{} on three complementary long-horizon memory settings, each stressing a different evidence-selection challenge.
\textit{Text memory} is evaluated on LoCoMo~\citep{locomo2024}, where questions over long conversational histories require source-sensitive, temporal, multi-hop, single-hop, open-domain, and adversarial evidence. We report results over all \(1{,}986\) questions.
\textit{Visual memory} is evaluated on EMemBench Visual Games~\citep{emembench2026}, where answers depend on frame-level observations, object states, event occurrences, and visual state changes.
\textit{Agent-use memory} is evaluated on AMA-Bench~\citep{amabench2026}, where questions involve action traces, causal dependencies, state updates, and abstract task states in agent trajectories.


\paragraph{Baselines and provenance.}
We compare against the strongest available baselines reported for each benchmark.
Because these systems were originally developed under different readers, prompts, and deployment assumptions, we distinguish benchmark-native results from shared-interface or adapted comparisons when applicable.
Our interpretation therefore focuses on the score--token trade-off under the reported evaluation protocol, rather than claiming that every adapted baseline exactly reproduces the original full system.

\begin{table*}[t]
\centering

\resizebox{\textwidth}{!}{%
\begin{tabular}{l cc cc cc cc cc cc r}
\toprule
& \multicolumn{2}{c}{\textbf{MultiHop} ($N{=}282$)}
& \multicolumn{2}{c}{\textbf{Temporal} ($N{=}321$)}
& \multicolumn{2}{c}{\textbf{OpenDomain} ($N{=}96$)}
& \multicolumn{2}{c}{\textbf{SingleHop} ($N{=}841$)}
& \multicolumn{2}{c}{\textbf{Adversarial} ($N{=}446$)}
& \multicolumn{2}{c}{\textbf{Overall} ($N{=}1986$)}
& \\
\cmidrule(lr){2-3}\cmidrule(lr){4-5}\cmidrule(lr){6-7}\cmidrule(lr){8-9}\cmidrule(lr){10-11}\cmidrule(lr){12-13}
\textbf{Method} & F1 & BLEU & F1 & BLEU & F1 & BLEU & F1 & BLEU & F1 & BLEU & F1 & BLEU & \textbf{Tokens} \\
\midrule

LoCoMo~\citep{locomo2024} & 0.28 & 0.18 & 0.09 & 0.06 & 0.16 & 0.15 & \textbf{0.62} & \textbf{0.54} & \textbf{0.53} & \textbf{0.51} & \underline{0.44} & \underline{0.39} & 16,910 \\
ReadAgent~\citep{lee2024readagent} & 0.15 & 0.10 & 0.04 & 0.03 & 0.09 & 0.08 & 0.12 & 0.10 & 0.07 & 0.06 & 0.10 & 0.08 & 805 \\
MemoryBank~\citep{zhong2024memorybank} & 0.06 & 0.05 & 0.02 & 0.02 & 0.06 & 0.05 & 0.08 & 0.07 & 0.04 & 0.04 & 0.06 & 0.05 & 569 \\
MemGPT~\citep{packer2023memgpt} & 0.30 & 0.23 & 0.17 & 0.13 & 0.12 & 0.12 & 0.60 & \underline{0.53} & 0.35 & 0.34 & 0.41 & 0.36 & 16,987 \\
A-Mem~\citep{amem2025agentic} & \underline{0.33} & \textbf{0.24} & \underline{0.39} & \underline{0.31} & \underline{0.17} & \underline{0.16} & 0.48 & 0.43 & 0.36 & 0.36 & 0.41 & 0.35 & 1,216 \\
\textbf{\method{}} & \textbf{0.33} & \underline{0.23} & \textbf{0.46} & \textbf{0.40} & \textbf{0.21} & \textbf{0.18} & \underline{0.61} & 0.50 & \underline{0.39} & \underline{0.36} & \textbf{0.48} & \textbf{0.40} & 1,073 \\
\bottomrule
\end{tabular}%
}
\caption{QA performance on the LoCoMo benchmark with GPT-4o. Overall scores are weighted by the number of questions per category ($N{=}1986$). Best results in \textbf{bold}, second-best \underline{underlined}.}
\label{tab:locomo-results}
\end{table*}

\paragraph{Metrics.} For \textit{answer quality}, on LoCoMo, we report F1 and BLEU for each question category and compute weighted overall scores according to the number of questions in each category.
On EMemBench Visual Games, we report accuracy, F1, and average evidence tokens per question.
On AMA-Bench, we report accuracy on Recall, Causal Inference, State Updating, and State Abstraction, together with their average.

For \textit{evidence-token accounting}, following \citep{amem2025agentic},
we report tok/q as the average number of evidence tokens exposed to the answer-time reader per question.
This measures the size of the memory interface rather than total API usage.

\paragraph{Implementation details.} In \method{},
the Structured Memory Writer and the answer-time reader are both fixed; the only
trained component is the lightweight Evidence Router. The supervision of the Route is
constructed offline from trajectory-QA data built over four interactive
environments: \crafter{}~\citep{hafner2022benchmarking},
\jericho{}~\citep{hausknecht2020interactive},
\sciworld{}~\citep{wang2022scienceworld}, and
\alfworld{}~\citep{shridhar2021alfworld}. We use GPT-5~\citep{openai2025gpt5} only as an offline
teacher to construct trajectory-grounded questions, answers, metadata, and weak
evidence annotations. Evidence steps are mapped to support-unit labels, question
metadata is used to derive query-anchor labels, and ordered multi-step evidence
is converted into anchor--support edge labels. These labels supervise the three
router heads for support, anchor, and edge scoring. Despite this weak and offline supervision, the trained router transfers to downstream benchmarks without benchmark-specific fine-tuning, suggesting that the structured support, anchor, and edge targets provide useful supervision for evidence selection.

In all main benchmark evaluations, the router is fixed and GPT-4o~\citep{openai2024gpt4o} is used as the
answer-time reader. Inference does not use target benchmark answers, gold
evidence, judge labels, benchmark scores, or reader outputs. We provide the data
construction protocol, router targets, training configuration, and leakage-control
rules in Appendix~\ref{app:weak_supervision} and Appendix~\ref{app:protocol}.



\subsection{Main Results}
\label{sec:main_results}
With the same Evidence Route, we evaluate our \method{} on text, visual, and agent-use memory tasks, demonstrating its effectiveness and generalization.

\paragraph{Text memory.} To evaluate the effectiveness of \method{} on long-horizon textual histories, we conduct experiments on LoCoMo. As shown in Table~\ref{tab:locomo-results}, \method{} achieves the best overall result, with \textbf{\(0.48\)} F1 and \textbf{\(0.40\)} BLEU, outperforming SOTA methods, \emph{e.g.}, LoCoMo~\citep{locomo2024}, A-Mem~\citep{amem2025agentic}, and MemGPT~\citep{memgpt}. Additionally, \method{} shows extreme token efficiency, consuming only \(1{,}073\) evidence tokens per question.
Compared with the LoCoMo reference (\(16{,}910\) tokens), \method{} improves answer quality while reducing the evidence context by about \(15.8\times\).
Compared with A-Mem, the strongest compact baseline in the table, \method{} improves \textbf{+0.07} overall F1 and \textbf{0.05} overall BLEU, while using slightly fewer tokens.
The gains are most pronounced on temporal and open-domain questions, suggesting that structured event-state evidence is useful for locating time-sensitive and source-sensitive support.
However, \method{} remains below the LoCoMo reference on adversarial questions, indicating that compact evidence selection can still miss distractor-sensitive support.


\begin{table*}[t]
\centering

\small
\setlength{\tabcolsep}{3pt}
\begin{tabular}{@{}l c c c c c r@{}}
\toprule
\textbf{Method} & \textbf{Recall} & \textbf{Causal Inf.} & \textbf{State Updating} & \textbf{State Abstraction} & \textbf{Average} & \textbf{tok/q} \\
\midrule
AriGraph~\cite{arigraph2024}
& \textbf{0.47} 
& \textbf{0.58} 
& 0.36 
& \textbf{0.29} 
& \textbf{0.43} 
& 2,207 \\

HippoRAG2~\cite{gutierrez2025hipporag2} 
& \underline{0.46} 
& 0.51 
& \textbf{0.39} 
& \underline{0.24} 
& \underline{0.41} 
& 3,633 \\

A-Mem~\cite{amem2025agentic} 
& 0.42 
& 0.54 
& \underline{0.38} 
& 0.19 
& 0.40 
& 2,579 \\

GraphRAG~\cite{edge2024graphrag} 
& 0.39 
& 0.47 
& 0.36 
& 0.17 
& 0.36 
& \underline{1,274} \\

mem0~\cite{chhikara2025mem0} 
& 0.32 
& \underline{0.56} 
& 0.32 
& 0.19 
& 0.35 
& 20,932 \\

\textbf{\method{}} 
& 0.36 
& \textbf{0.58} 
& 0.36 
& 0.21 
& 0.39 
& \textbf{735} \\
\bottomrule
\end{tabular}
\caption{AMA-Bench results. Best results in \textbf{bold}, second-best \underline{underlined}.}
\label{tab:amabench-results}
\end{table*}



\paragraph{Visual memory.} 
On EMemBench Visual Games, \method{} proves that retrieving specific occurrences is far more effective than exposing broad context. Specifically, as illustrated in Table~\ref{tab:memory-env-results}, \method{} obtains the best F1 (\(0.39\)) and the second-best accuracy (\(0.36\)), while using only \(189\) evidence tokens per question. While
A-Mem~\citep{amem2025agentic} achieves the highest accuracy (\(0.42\)), it requires \(6{,}145\) tokens per question, more than \(32\times\) the evidence budget of \method{}.
Similarly, Graph-style methods, \emph{e.g.}, HippoRAG2~\citep{gutierrez2025hipporag2} and AriGraph~\citep{arigraph2024}, can also obtain comparable accuracy, but cost substantially larger evidence contexts.
We conjecture that visual memory questions heavily depend on localized observations, object states, and event occurrences. By organizing visual memory into compact, structured packs, \method{} avoids the noise that plagues pure retrieval or massive context windows, reaching better per-token evidence. 

\paragraph{Agent-use memory.} We also verify the effectiveness of \method{} on the agent-use memory task using AMA-Bench. As shown in Table~\ref{tab:amabench-results}, while explicit graph traversal methods hold an edge in global trajectory tasks, \method{} provides the most efficient paradigm. Concretely, compared with AriGraph~\citep{arigraph2024} (2,207 tokens) and HippoRAG2~\citep{gutierrez2025hipporag2} (3,633 tokens), \method{} costs an extremely tight budget of only \textbf{735} tokens while achieves competitive average score. 
Notably, \method{} ties the best causal-inference score (\(0.58\)).  This demonstrates that its anchor-support evidence chains successfully capture local causal dependencies between actions and states.
At the same time, \method{} is weaker on Recall and State Abstraction compared with graph-based methods. This suggests that while structured routing excels at extracting specific causal links, agent-use memory still benefits from broader trajectory coverage or explicit graph traversal.

\paragraph{Discussion.} In summary, these results delineate a clear operational boundary for \method{}. Rather than attempting to universally dominate raw accuracy through context expansion, \method{} establishes a highly efficient score--token Pareto frontier. It excels in precision-driven scenarios, where answers hinge on granular events, explicit state transitions, temporal links, or specific provenance. When facing scenarios requiring holistic trajectory recall or macro-level abstraction, which naturally benefit from massive context windows or exhaustive graph traversals, \method{} can also achieve competitive results with much fewer token cost.

\begin{table}[t]
\centering

\small
\setlength{\tabcolsep}{4pt}
\begin{tabular}{@{}l c c r@{}}
\toprule
\textbf{Method} & \textbf{Acc} & \textbf{F1} & \textbf{tok/q} \\
\midrule
A-Mem~\cite{amem2025agentic} & \textbf{0.42} & 0.32  & 6,145 \\
mem0~\cite{chhikara2025mem0} & 0.33 & 0.09  & 2,752 \\
HippoRAG2~\cite{gutierrez2025hipporag2} & 0.34 & \underline{0.36}  & 2,938 \\
AriGraph~\cite{arigraph2024} & 0.34 & 0.34 & 1,293 \\
GraphRAG~\cite{edge2024graphrag} & 0.12 & 0.12  & \underline{779} \\
\textbf{\method{}} & \underline{0.36} & \textbf{0.39}  & \textbf{189} \\
\bottomrule
\end{tabular}
\caption{Results on EMemBench Visual Games. Best results are shown in \textbf{bold} and second-best results are \underline{underlined}. tok/q denotes the average number of evidence tokens exposed to the reader per question.}
\label{tab:memory-env-results}
\end{table}

\begin{table}[t]
\centering

\small
\setlength{\tabcolsep}{4pt}
\begin{tabular}{@{}l l c r@{}}
\toprule
\textbf{Variant} & \textbf{Representation} & \textbf{Acc} & \textbf{tok/q} \\
\midrule
\method{} 
& Full unit
& \textbf{0.3495} 
& 454 \\

Plain chunk 
& Text chunk
& 0.2517 
& 392 \\

Event-only 
& Event field
& 0.1163 
& 167 \\

Object-only 
& Object field
& 0.0953 
& 332 \\
\bottomrule
\end{tabular}
\caption{
Representation ablation on EMemBench External.
The full structured unit outperforms plain chunks and single-field variants.
}
\vspace{-1em}
\label{tab:representation-memory-unit-ablation}
\end{table}

\newcommand{\est}[1]{\textit{#1}$^{\ast}$}

\begin{table}[t]
\centering
\small

\setlength{\tabcolsep}{2pt}
\resizebox{\linewidth}{!}{
\begin{tabular}{lcccc}
\toprule
\textbf{Reader model}
& \textbf{GraphRAG}
& \textbf{LightMem}
& \textbf{HippoRAG}
& \textbf{\method{}} \\
\midrule
DeepSeek-v4-pro~\cite{deepseekai2026deepseekv4}
& 35.1
& 31.0
& 47.6
& \textbf{49.6} \\
Gemini3Flash~\cite{googledeepmind2025gemini3flash}
& 34.3
& 33.0
& 38.1
& \textbf{40.1} \\
GLM5.1~\cite{glm5team2026glm5vibecodingagentic}
& 42.5
& 48.8
& \textbf{53.0}
& 49.6 \\
\bottomrule
\end{tabular}}
\caption{
Reader-model sensitivity on EMemBench text external.
Each cell reports the official score under the corresponding reader.
}
\label{tab:reader_sensitivity_app}
\end{table}

\subsection{Ablation Study}
\label{sec:ablation}

We conduct ablations on EMemBench to isolate how specific components of \method{} contribute to the
score--token trade-off. Unless otherwise stated, ablations use the same fixed
reader and the same evaluation protocol as the corresponding benchmark setting.
We systematically study three questions: (i) \textbf{Representation}: Are structured scene--event units more superior than unstructured text chunks? (ii) \textbf{Routing mechanisms}: Whether router components
such as anchors, support-chain expansion, and compact packing improve evidence
selection? (iii) \textbf{Reader Agnosticism}: Do these gains generalize across different LLM readers? 

\paragraph{Structured Units vs. Plain Chunks:} 
Replacing the full structured units with standard unstructured text chunks causes a severe accuracy drop \(0.3495\) to \(0.2517\) (Table~\ref{tab:representation-memory-unit-ablation}), despite both consuming a comparable token budget.
This validates that the advantage of \method{} lies not in merely compressing text, but in explicitly isolating searchable attributes (events, objects, and relations) that standard dense retrieval often conflates into noisy context.

\paragraph{Failure of Single-Filed Compression:} We further ablate the units into extreme single-field variants (Event-only and Object-only). While Event-only uses the fewest tokens (167), its accuracy degrades catastrophically to about 0.1163 and 0.0953, respectively. This proves a core philosophy of \method{}: compactness alone is meaningless if the evidence ceases to be answer-sufficient.
Answering long-horizon questions intrinsically requires the synergy of the entire \textit{which--what--where--how} schema:
events identify occurrences, objects identify entities, relations connect them, and timeline fields preserve state changes.
Stripping away any of these dimensions breaks the logical chain required by the reader.

\paragraph{Plug-and-Play Reader agnosticism.} 
Finally, we investigate whether the compact evidence packs generated by \method{} are universally digestible by different LLMs, \emph{i.e.}, DeepSeek-V4-Pro~\citep{deepseekai2026deepseekv4},
Gemini 3 Flash~\citep{googledeepmind2025gemini3flash},
and GLM-5.1~\citep{glm5team2026glm5vibecodingagentic}
on the EMemBench text external subset.
As detailed in Table~\ref{tab:reader_sensitivity_app},
\method{} is best or near-best under different readers, while using fewer
evidence tokens than HippoRAG. This suggests that the observed trend is not
specific to one answer-time model, although the absolute score still depends on
reader strength.

\section{Conclusion}

We introduced \method{}, a structured evidence-routing interface for long-horizon memory question answering. It converts heterogeneous histories into scene--event units and selects compact evidence packs for a fixed reader. Across text, visual, and agent-use memory benchmarks, \method{} achieves a strong score--token trade-off, especially when answers depend on localized events, states, temporal relations, or provenance evidence. Its weaker performance on recall-heavy and abstraction-heavy agent-use questions highlights the limits of compact routing, suggesting that structured evidence routing is a practical but not universal interface for efficient long-horizon memory QA.



\section*{Limitations}
This work focuses on query-time evidence selection for long-horizon memory rather than end-to-end agent training.
Our claims are bounded to the evaluated benchmarks, readers, and evidence-token budgets reported in the experiments.
Structured routing can under-select evidence, miss support steps in multi-hop questions, or fail when benchmark assumptions diverge from trajectory-style training data.

Additional failure modes and claim boundaries are discussed in Appendix~\ref{app:limitations}.

\section*{Ethical Considerations and Artifact Use}
This work uses existing public research benchmarks and benchmark-style interactive trajectories for long-horizon memory question answering. We use these artifacts for research evaluation only and follow their released terms, licenses, and intended benchmark protocols where available. We do not redistribute original benchmark data beyond the permitted research-use setting.

The weak-supervision data constructed for router training is derived from interactive environments and trajectory-based QA examples, rather than newly collected human-subject data. We check generated examples and released artifacts to avoid including names, uniquely identifying personal information, or offensive content. Artifacts created in this work, including structured memory units, weak router labels, and evaluation outputs, are intended for research on memory-side evidence selection and should not be used for non-research deployment without additional privacy, safety, and data-governance review.

\bibliography{references}

@inproceedings{lewis2020retrieval,
  title={Retrieval-Augmented Generation for Knowledge-Intensive {NLP} Tasks},
  author={Lewis, Patrick and Perez, Ethan and Piktus, Aleksandra and Petroni, Fabio and Karpukhin, Vladimir and Goyal, Naman and K{\"u}ttler, Heinrich and Lewis, Mike and Yih, Wen-tau and Rockt{\"a}schel, Tim and Riedel, Sebastian and Kiela, Douwe},
  booktitle={Advances in Neural Information Processing Systems},
  volume={33},
  year={2020}
}

@inproceedings{guu2020realm,
  title={{REALM}: Retrieval-Augmented Language Model Pre-Training},
  author={Guu, Kelvin and Lee, Kenton and Tung, Zora and Pasupat, Panupong and Chang, Mingwei},
  booktitle={International Conference on Machine Learning},
  year={2020}
}

@article{izacard2023atlas,
  title={Atlas: Few-shot Learning with Retrieval Augmented Language Models},
  author={Izacard, Gautier and Lewis, Patrick and Lomeli, Maria and Hosseini, Lucas and Petroni, Fabio and Schick, Timo and Dwivedi-Yu, Jane and Joulin, Armand and Riedel, Sebastian and Grave, Edouard},
  journal={Journal of Machine Learning Research},
  volume={24},
  number={251},
  pages={1--43},
  year={2023}
}

@inproceedings{he2024gretriever,
  title={{G-Retriever}: Retrieval-Augmented Generation for Textual Graph Understanding and Question Answering},
  author={He, Xiaoxin and Tian, Yijun and Sun, Yifei and Chawla, Nitesh V. and Laurent, Thomas and LeCun, Yann and Bresson, Xavier and Hooi, Bryan},
  booktitle={Advances in Neural Information Processing Systems},
  year={2024}
}

@inproceedings{park2023generative,
  title={Generative Agents: Interactive Simulacra of Human Behavior},
  author={Park, Joon Sung and O'Brien, Joseph C. and Cai, Carrie J. and Morris, Meredith Ringel and Liang, Percy and Bernstein, Michael S.},
  booktitle={ACM Symposium on User Interface Software and Technology},
  year={2023}
}

@inproceedings{shinn2023reflexion,
  title={Reflexion: Language Agents with Verbal Reinforcement Learning},
  author={Shinn, Noah and Cassano, Federico and Gopinath, Ashwin and Narasimhan, Karthik and Yao, Shunyu},
  booktitle={Advances in Neural Information Processing Systems},
  year={2023}
}

@article{packer2023memgpt,
  title={{MemGPT}: Towards {LLMs} as Operating Systems},
  author={Packer, Charles and Wooders, Sarah and Lin, Kevin and Fang, Vivian and Patil, Shishir G. and Stoica, Ion and Gonzalez, Joseph E.},
  journal={arXiv preprint arXiv:2310.08560},
  year={2023}
}

@article{wang2023voyager,
  title={Voyager: An Open-Ended Embodied Agent with Large Language Models},
  author={Wang, Guanzhi and Xie, Yuqi and Jiang, Yunfan and Mandlekar, Ajay and Xiao, Chaowei and Zhu, Yuke and Fan, Linxi and Anandkumar, Anima},
  journal={arXiv preprint arXiv:2305.16291},
  year={2023}
}

@article{tulving1972episodic,
  title={Episodic and Semantic Memory},
  author={Tulving, Endel},
  journal={Organization of Memory},
  pages={381--403},
  year={1972},
  publisher={Academic Press}
}

@inproceedings{nuxoll2007extending,
  title={Extending Cognitive Architecture with Episodic Memory},
  author={Nuxoll, Andrew M. and Laird, John E.},
  booktitle={AAAI Conference on Artificial Intelligence},
  year={2007}
}

@inproceedings{blundell2016model,
  title={Model-Free Episodic Control},
  author={Blundell, Charles and Uria, Benigno and Pritzel, Alexander and Li, Yazhe and Ruderman, Avraham and Leibo, Joel Z and Rae, Jack and Wierstra, Daan and Hassabis, Demis},
  booktitle={arXiv preprint arXiv:1606.04460},
  year={2016}
}

@inproceedings{pritzel2017neural,
  title={Neural Episodic Control},
  author={Pritzel, Alexander and Uria, Benigno and Srinivasan, Sriram and Puigdom{\`e}nech Badia, Adri{\`a} and Vinyals, Oriol and Hassabis, Demis and Wierstra, Daan and Blundell, Charles},
  booktitle={International Conference on Machine Learning},
  year={2017}
}

@article{amem2025agentic,
  title={{A-MEM}: Agentic Memory for {LLM} Agents},
  author={Xu, Wujiang and Liang, Zujie and Mei, Kai and Gao, Hang and Tan, Juntao and Zhang, Yongfeng},
  journal={arXiv preprint arXiv:2502.12110},
  year={2025}
}

@article{memoryos2025,
  title={Memory {OS} of {AI} Agent},
  author={Kang, Jiazheng and Ji, Mingming and Zhao, Zhe and Bai, Ting},
  journal={arXiv preprint arXiv:2506.06326},
  year={2025}
}

@article{seem2026structured,
  title={Structured Episodic Event Memory},
  author={Lu, Zhengxuan and Li, Dongfang and Shi, Yukun and Wang, Beilun and Wang, Longyue and Hu, Baotian},
  journal={arXiv preprint arXiv:2601.06411},
  year={2026}
}

@article{lightmem2025,
  title={{LightMem}: Lightweight and Efficient Memory-Augmented Generation},
  author={Fang, Jizhan and Deng, Xinle and Xu, Haoming and Jiang, Ziyan and Tang, Yuqi and Xu, Ziwen and Deng, Shumin and Yao, Yunzhi and Wang, Mengru and Qiao, Shuofei and Chen, Huajun and Zhang, Ningyu},
  journal={arXiv preprint arXiv:2510.18866},
  year={2025}
}

@article{arigraph2024,
  title={AriGraph: Learning Knowledge Graph World Models with Episodic Memory for {LLM} Agents},
  author={Anokhin, Petr and Semenov, Nikita and Sorokin, Artyom Y. and Evseev, Dmitry and Burtsev, Mikhail and Burnaev, Evgeny},
  journal={CoRR},
  volume={abs/2407.04363},
  year={2024}
}

@inproceedings{hafner2022benchmarking,
  title={Benchmarking the Spectrum of Agent Capabilities},
  author={Hafner, Danijar},
  booktitle={International Conference on Learning Representations},
  year={2022}
}

@inproceedings{hausknecht2020interactive,
  title={Interactive Fiction Games: A Colossal Adventure},
  author={Hausknecht, Matthew and Ammanabrolu, Prithviraj and C{\^o}t{\'e}, Marc-Alexandre and Yuan, Xingdi},
  booktitle={AAAI Conference on Artificial Intelligence},
  year={2020}
}

@inproceedings{wang2022scienceworld,
  title={{ScienceWorld}: Is your Agent Smarter than a 5th Grader?},
  author={Wang, Ruoyao and Jansen, Peter and C{\^o}t{\'e}, Marc-Alexandre and Ammanabrolu, Prithviraj},
  booktitle={Conference on Empirical Methods in Natural Language Processing},
  year={2022}
}

@inproceedings{shridhar2021alfworld,
  title={{ALFWorld}: Aligning Text and Embodied Environments for Interactive Learning},
  author={Shridhar, Mohit and Yuan, Xingdi and C{\^o}t{\'e}, Marc-Alexandre and Bisk, Yonatan and Trischler, Adam and Hausknecht, Matthew},
  booktitle={International Conference on Learning Representations},
  year={2021}
}

@article{locomo2024,
  title={Evaluating Very Long-Term Conversational Memory of {LLM} Agents},
  author={Maharana, Adyasha and Lee, Dong-Ho and Tulyakov, Sergey and Bansal, Mohit and Barbieri, Francesco and Fang, Yuwei},
  journal={arXiv preprint arXiv:2402.17753},
  year={2024}
}

@article{emembench2026,
  title={{EMemBench}: Interactive Benchmarking of Episodic Memory for {VLM} Agents},
  author={Li, Xinze and Zhu, Ziyue and Liu, Siyuan and Ma, Yubo and Zang, Yuhang and Cao, Yixin and Sun, Aixin},
  journal={arXiv preprint arXiv:2601.16690},
  year={2026}
}

@article{amabench2026,
  title={{AMA-Bench}: Evaluating Long-Horizon Memory for Agentic Applications},
  author={Zhao, Yujie and Yuan, Boqin and Huang, Junbo and Yuan, Haocheng and Yu, Zhongming and Xu, Haozhou and Hu, Lanxiang and Shankarampeta, Abhilash and Huang, Zimeng and Ni, Wentao and Tian, Yuandong and Zhao, Jishen},
  journal={arXiv preprint arXiv:2602.22769},
  year={2026}
}

@inproceedings{chen2017reading,
  title = {Reading Wikipedia to Answer Open-Domain Questions},
  author = {Chen, Danqi and Fisch, Adam and Weston, Jason and Bordes, Antoine},
  booktitle = {Proceedings of the 55th Annual Meeting of the Association for Computational Linguistics (Volume 1: Long Papers)},
  pages = {1870--1879},
  year = {2017},
  publisher = {Association for Computational Linguistics},
  doi = {10.18653/v1/P17-1171},
  url = {https://aclanthology.org/P17-1171}
}

@inproceedings{karpukhin2020dense,
  title = {Dense Passage Retrieval for Open-Domain Question Answering},
  author = {Karpukhin, Vladimir and Oguz, Barlas and Min, Sewon and Lewis, Patrick and Wu, Ledell and Edunov, Sergey and Chen, Danqi and Yih, Wen-tau},
  booktitle = {Proceedings of the 2020 Conference on Empirical Methods in Natural Language Processing},
  pages = {6769--6781},
  year = {2020}
}

@inproceedings{jimenez2024swebench,
  title = {{SWE}-bench: Can Language Models Resolve Real-world Github Issues?},
  author = {Jimenez, Carlos E. and Yang, John and Wettig, Alexander and Yao, Shunyu and Pei, Kexin and Press, Ofir and Narasimhan, Karthik R.},
  booktitle = {The Twelfth International Conference on Learning Representations},
  year = {2024},
  url = {https://openreview.net/forum?id=VTF8yNQM66}
}

@inproceedings{yang2024sweagent,
  title = {{SWE}-agent: Agent-Computer Interfaces Enable Automated Software Engineering},
  author = {Yang, John and Jimenez, Carlos E. and Wettig, Alexander and Lieret, Kilian and Yao, Shunyu and Narasimhan, Karthik R. and Press, Ofir},
  booktitle = {The Thirty-eighth Annual Conference on Neural Information Processing Systems},
  year = {2024},
  url = {https://arxiv.org/abs/2405.15793}
}

@inproceedings{qin2024toolllm,
  title = {{ToolLLM}: Facilitating Large Language Models to Master 16000+ Real-world APIs},
  author = {Qin, Yujia and Liang, Shihao and Ye, Yining and Zhu, Kunlun and Yan, Lan and Lu, Yaxi and Lin, Yankai and Cong, Xin and Tang, Xiangru and Qian, Bill and Zhao, Sihan and Hong, Lauren and Tian, Runchu and Xie, Ruobing and Zhou, Jie and Gerstein, Mark and Li, Dahai and Liu, Zhiyuan and Sun, Maosong},
  booktitle = {The Twelfth International Conference on Learning Representations},
  year = {2024},
  url = {https://openreview.net/forum?id=dHng2O0Jjr}
}

@inproceedings{liu2024agentbench,
  title = {{AgentBench}: Evaluating {LLMs} as Agents},
  author = {Liu, Xiao and Yu, Hao and Zhang, Hanchen and Xu, Yifan and Lei, Xuanyu and Lai, Hanyu and Gu, Yu and Ding, Hangliang and Men, Kaiwen and Yang, Kejuan and Zhang, Shudan and Deng, Xiang and Zeng, Aohan and Du, Zhengxiao and Zhang, Chenhui and Shen, Sheng and Zhang, Tianjun and Su, Yu and Sun, Huan and Huang, Minlie and Dong, Yuxiao and Tang, Jie},
  booktitle = {The Twelfth International Conference on Learning Representations},
  year = {2024},
  url = {https://openreview.net/forum?id=zAdUB0aCTQ}
}

@inproceedings{koh2024visualwebarena,
  title = {{VisualWebArena}: Evaluating Multimodal Agents on Realistic Visual Web Tasks},
  author = {Koh, Jing Yu and Lo, Robert and Jang, Lawrence and Duvvur, Vikram and Lim, Ming and Huang, Po-Yu and Neubig, Graham and Zhou, Shuyan and Salakhutdinov, Ruslan and Fried, Daniel},
  booktitle = {Proceedings of the 62nd Annual Meeting of the Association for Computational Linguistics},
  pages = {881--905},
  year = {2024},
  publisher = {Association for Computational Linguistics},
  doi = {10.18653/v1/2024.acl-long.50},
  url = {https://aclanthology.org/2024.acl-long.50/}
}

@article{xie2024osworld,
  title = {{OSWorld}: Benchmarking Multimodal Agents for Open-Ended Tasks in Real Computer Environments},
  author = {Xie, Tianbao and Zhang, Danyang and Chen, Jixuan and Li, Xiaochuan and Zhao, Siheng and Cao, Ruisheng and Hua, Toh Jing and Cheng, Zhoujun and Shin, Dongchan and Lei, Fangyu and Liu, Yitao and Xu, Yiheng and Zhou, Shuyan and Savarese, Silvio and Xiong, Caiming and Zhong, Victor and Yu, Tao},
  journal = {arXiv preprint arXiv:2404.07972},
  year = {2024},
  url = {https://arxiv.org/abs/2404.07972}
}

@article{li2024personalllmagents,
  title = {Personal {LLM} Agents: Insights and Survey about the Capability, Efficiency and Security},
  author = {Li, Yuanchun and Wen, Hao and Wang, Weijun and Li, Xiangyu and Yuan, Yizhen and Liu, Guohong and Liu, Jiacheng and Xu, Wenxing and Wang, Xiang and Sun, Yi and Kong, Rui and Wang, Yile and Geng, Hanfei and Luan, Jian and Jin, Xuefeng and Ye, Zilong and Xiong, Guanjing and Zhang, Fan and Li, Xiang and Xu, Mengwei and Li, Zhijun and Li, Peng and Liu, Yang and Zhang, Ya-Qin and Liu, Yunxin},
  journal = {arXiv preprint arXiv:2401.05459},
  year = {2024},
  url = {https://arxiv.org/abs/2401.05459}
}

@inproceedings{zhong2024memorybank,
  title = {{MemoryBank}: Enhancing Large Language Models with Long-Term Memory},
  author = {Zhong, Wanjun and Guo, Lianghong and Gao, Qiqi and Ye, He and Wang, Yanlin},
  booktitle = {Proceedings of the AAAI Conference on Artificial Intelligence},
  volume = {38},
  number = {17},
  pages = {19724--19731},
  year = {2024},
  doi = {10.1609/aaai.v38i17.29946},
  url = {https://ojs.aaai.org/index.php/AAAI/article/view/29946}
}

@article{chhikara2025mem0,
  title={Mem0: Building production-ready ai agents with scalable long-term memory},
  author={Chhikara, Prateek and Khant, Dev and Aryan, Saket and Singh, Taranjeet and Yadav, Deshraj},
  journal={arXiv preprint arXiv:2504.19413},
  year={2025}
}

@inproceedings{lee2024readagent,
  title     = {A Human-Inspired Reading Agent with Gist Memory of Very Long Contexts},
  author    = {Lee, Kuang-Huei and Chen, Xinyun and Furuta, Hiroki and Canny, John and Fischer, Ian},
  booktitle = {Proceedings of the 41st International Conference on Machine Learning},
  pages     = {26396--26415},
  year      = {2024},
  volume    = {235},
  series    = {Proceedings of Machine Learning Research},
  publisher = {PMLR},
  url       = {https://proceedings.mlr.press/v235/lee24c.html}
}

@article{edge2024graphrag,
  title   = {From Local to Global: A Graph RAG Approach to Query-Focused Summarization},
  author  = {Edge, Darren and Trinh, Ha and Cheng, Newman and Bradley, Joshua and Chao, Alex and Mody, Apurva and Truitt, Steven and Metropolitansky, Dasha and Ness, Robert Osazuwa and Larson, Jonathan},
  journal = {arXiv preprint arXiv:2404.16130},
  year    = {2024},
  url     = {https://arxiv.org/abs/2404.16130}
}

@inproceedings{gutierrez2025hipporag2,
  title     = {From {RAG} to Memory: Non-Parametric Continual Learning for Large Language Models},
  author    = {Guti{\'e}rrez, Bernal Jim{\'e}nez and Shu, Yiheng and Qi, Weijian and Zhou, Sizhe and Su, Yu},
  booktitle = {Proceedings of the 42nd International Conference on Machine Learning},
  pages     = {21497--21515},
  year      = {2025},
  volume    = {267},
  series    = {Proceedings of Machine Learning Research},
  publisher = {PMLR},
  url       = {https://proceedings.mlr.press/v267/gutierrez25a.html}
}

@misc{deepseekai2026deepseekv4,
  title  = {DeepSeek-V4: Towards Highly Efficient Million-Token Context Intelligence},
  author = {{DeepSeek-AI}},
  year   = {2026},
  url    = {https://huggingface.co/deepseek-ai/DeepSeek-V4-Pro}
}

@misc{googledeepmind2025gemini3flash,
  title        = {Gemini 3 Flash Model Card},
  author       = {{Google DeepMind}},
  year         = {2025},
  month        = dec,
  howpublished = {Model card},
  url          = {https://storage.googleapis.com/deepmind-media/Model-Cards/Gemini-3-Flash-Model-Card.pdf}
}

@misc{glm5team2026glm5vibecodingagentic,
  title         = {{GLM}-5: from Vibe Coding to Agentic Engineering},
  author        = {{GLM-5-Team} and Zeng, Aohan and Lv, Xin and Hou, Zhenyu and Du, Zhengxiao and Zheng, Qinkai and Chen, Bin and Yin, Da and Ge, Chendi and Huang, Chenghua and Xie, Chengxing and Zhu, Chenzheng and Yin, Congfeng and Wang, Cunxiang and Pan, Gengzheng and Zeng, Hao and Zhang, Haoke and Wang, Haoran and Chen, Huilong and Zhang, Jiajie and Jiao, Jian and Guo, Jiaqi and Wang, Jingsen and Du, Jingzhao and Wu, Jinzhu and Wang, Kedong and Li, Lei and Fan, Lin and Zhong, Lucen and Liu, Mingdao and Zhao, Mingming and Du, Pengfan and Dong, Qian and Lu, Rui and Li, Shuang and Cao, Shulin and Liu, Song and Jiang, Ting and Chen, Xiaodong and Zhang, Xiaohan and Huang, Xuancheng and Dong, Xuezhen and Xu, Yabo and Wei, Yao and An, Yifan and Niu, Yilin and Zhu, Yitong and Wen, Yuanhao and Cen, Yukuo and Bai, Yushi and Qiao, Zhongpei and Wang, Zihan and Wang, Zikang and Zhu, Zilin and Liu, Ziqiang and Li, Zixuan and others},
  year          = {2026},
  eprint        = {2602.15763},
  archivePrefix = {arXiv},
  primaryClass  = {cs.LG},
  url           = {https://arxiv.org/abs/2602.15763}
}

@misc{openai2025gpt5,
  title        = {Introducing {GPT}-5},
  author       = {{OpenAI}},
  year         = {2025},
  month        = aug,
  howpublished = {OpenAI blog},
  url          = {https://openai.com/index/introducing-gpt-5/}
}

@misc{openai2024gpt4o,
  title        = {{GPT}-4o System Card},
  author       = {{OpenAI}},
  year         = {2024},
  month        = aug,
  howpublished = {System card},
  url          = {https://openai.com/index/gpt-4o-system-card/}
}

@inproceedings{bai2024longbench,
  title = "{L}ong{B}ench: A Bilingual, Multitask Benchmark for Long Context Understanding",
  author = "Bai, Yushi and
    Lv, Xin and
    Zhang, Jiajie and
    Lyu, Hongchang and
    Tang, Jiankai and
    Huang, Zhidian and
    Du, Zhengxiao and
    Liu, Xiao and
    Zeng, Aohan and
    Hou, Lei and
    Dong, Yuxiao and
    Tang, Jie and
    Li, Juanzi",
  editor = "Ku, Lun-Wei and
    Martins, Andre and
    Srikumar, Vivek",
  booktitle = "Proceedings of the 62nd Annual Meeting of the Association for Computational Linguistics (Volume 1: Long Papers)",
  month = aug,
  year = "2024",
  address = "Bangkok, Thailand",
  publisher = "Association for Computational Linguistics",
  pages = "3119--3137",
  url = "https://aclanthology.org/2024.acl-long.172/",
  doi = "10.18653/v1/2024.acl-long.172"
}

@article{liu2024lost,
  title = "Lost in the Middle: How Language Models Use Long Contexts",
  author = "Liu, Nelson F. and
    Lin, Kevin and
    Hewitt, John and
    Paranjape, Ashwin and
    Bevilacqua, Michele and
    Petroni, Fabio and
    Liang, Percy",
  journal = "Transactions of the Association for Computational Linguistics",
  volume = "12",
  year = "2024",
  address = "Cambridge, MA",
  publisher = "MIT Press",
  pages = "157--173",
  url = "https://aclanthology.org/2024.tacl-1.9/",
  doi = "10.1162/tacl_a_00638"
}

@article{bai2025qwen25vl,
  title = {Qwen2.5-VL Technical Report},
  author = {Bai, Shuai and Chen, Keqin and Liu, Xuejing and Wang, Jialin and Ge, Wenbin and Song, Sibo and Dang, Kai and Wang, Peng and Wang, Shijie and Tang, Jun and others},
  journal = {arXiv preprint arXiv:2502.13923},
  year = {2025},
  url = {https://arxiv.org/abs/2502.13923}
}

@article{yang2024qwen2,
  title = {Qwen2 Technical Report},
  author = {Yang, An and Yang, Baosong and Zhang, Binyuan and Hui, Bofei and Zheng, Bo and Yu, Bowen and Li, Chengyuan and Liu, Dayiheng and Huang, Fei and Wei, Haoran and others},
  journal = {arXiv preprint arXiv:2407.10671},
  year = {2024},
  url = {https://arxiv.org/abs/2407.10671}
}

\definecolor{unitback}{HTML}{F7F8FB}
\definecolor{unitframe}{HTML}{8A97B8}

\newtcblisting{unitlisting}[1]{
  enhanced,
  breakable,
  listing only,
  colback=unitback,
  colframe=unitframe,
  boxrule=0.6pt,
  arc=1.5mm,
  left=2mm,
  right=2mm,
  top=1.5mm,
  bottom=1.5mm,
  title={#1},
  fonttitle=\bfseries\small,
  coltitle=black,
  listing options={
    basicstyle=\ttfamily\footnotesize,
    breaklines=true,
    breakatwhitespace=true,
    columns=fullflexible,
    keepspaces=true
  }
}

\clearpage
\appendix


\addcontentsline{toc}{section}{Appendix Roadmap}
\label{app:roadmap}

\begin{table}[t]
\centering
\small
\setlength{\tabcolsep}{4pt}
\renewcommand{\arraystretch}{1.1}

\begin{tabularx}{\columnwidth}{@{} l X @{}}
\toprule
\textbf{Appendix Section} & \textbf{Content Description} \\
\midrule
\textbf{App.~\ref{app:method_details}} 
& Implementation details for structured scene--event units, candidate proposal, router scoring, and evidence packing. \\
\addlinespace
\textbf{App.~\ref{app:router_training}} & Router implementation, feature construction, weak-label generation, candidate labeling, and a training-instance example. \\
\addlinespace
\textbf{App.~\ref{app:weak_supervision}} 
& Weak-label construction, router training targets, and training--inference separation. \\
\addlinespace

\textbf{App.~\ref{app:protocol}} 
& Train/test separation, no-gold inference boundary, target adaptation policy, and leakage-control rules. \\
\addlinespace

\textbf{App.~\ref{app:benchmark_conversion}} 
& Benchmark conversion, scoring protocols, token accounting, and baseline provenance. \\
\addlinespace

\textbf{App.~\ref{app:supp_results}} 
& Supplemental results and diagnostic analyses that clarify, but do not replace, the main benchmark tables. \\
\addlinespace

\textbf{App.~\ref{app:case_studies}} 
& Qualitative case studies illustrating when compact structured evidence helps. \\
\addlinespace

\textbf{App.~\ref{app:limitations}} 
& Failure modes and claim-boundary clarifications. \\
\addlinespace

\textbf{App.~\ref{app:writer_prompt}} 
& Prompt used by the frozen structured memory writer. \\

\bottomrule
\end{tabularx}

\caption{Overview of appendix sections.}
\label{tab:appendix_overview}
\end{table}

\section{Additional Method Details}
\label{app:method_details}
\label{app:implementation}

This section provides implementation details that are omitted from the main text for space.  The evaluated component is the memory-side evidence interface: histories are written into structured scene--event units, candidate units are proposed from several high-recall signals, router scores are computed over units and unit pairs, and an ordered evidence pack is rendered for a fixed reader.

\subsection{Training and Inference Boundary}

The router is trained offline from weakly supervised trajectory-QA data collected
from environments separate from the downstream benchmark test instances. Weak
evidence annotations are converted into support-unit, query-anchor, and
anchor--support edge labels. The writer and reader are fixed, and only the
lightweight router is trained.

At benchmark inference time, all router parameters are fixed. The system selects
evidence using only the question, written memory units, candidate features, and
pre-answer relations. It does not access benchmark answers, gold evidence, judge
labels, benchmark scores, or reader outputs. Benchmark-specific code is limited
to input conversion, prompt rendering, and official or benchmark-compatible
scoring.

\subsection{Structured Scene--Event Units}

Each raw trajectory step, visual observation, conversation record, document segment, or tool-use log is converted into one or more structured scene--event units.  We use the same field schema across text, visual, and agent-use memory:
\[
  m_i=(t_i,e_i,o_i,\ell_i,\Delta s_i,r_i,p_i,u_i).
\]
Here, \(t_i\) is a time, step, frame, or turn identifier; \(e_i\) is the event or observation; \(o_i\) stores objects or entities; \(\ell_i\) stores location or source context; \(\Delta s_i\) records state changes when available; \(r_i\) stores relations to other units; \(p_i\) stores provenance metadata; and \(u_i\) points to the raw record.  Some fields can be empty when the source does not contain that information.  The unit does not replace the raw history; it provides searchable fields while preserving a pointer to the original evidence.

\subsection{Candidate Proposal}

Given a question \(q\) and written memory \(M\), candidate proposal constructs a high-recall pool \(C(q,M)\).  Proposal channels include semantic retrieval, lexical overlap, query-anchor matching, temporal proximity, object or entity overlap, visual continuity, causal links, and source continuity.  These channels are used only to collect candidates.  They do not determine the final reader-visible evidence.

\subsection{Evidence Packing and Rendering}

The packer is deterministic at inference time and has no learned parameters.  It converts router scores into an ordered evidence pack.  For single-hop questions, it selects one or a small number of high-support units.  For multi-hop questions, it starts from a high-anchor unit and expands through high-scoring support links.  The packer orders selected units according to the question type: temporal questions follow time or step order, causal questions place antecedent events before effects, provenance-sensitive questions preserve source pointers, and counting or comparison questions keep repeated occurrences separate when needed.

The rendered prompt contains only the selected evidence pack and the question.  The harness does not retrieve additional records, rescore candidates, call an additional reasoning model, or access gold answers, gold evidence, judge labels, benchmark scores, or reader outputs.

\section{Router Implementation and Training Details}
\label{app:router_training}

This section provides additional details on the router implementation, feature
construction, weak-label generation, candidate labeling, and training examples
used to train the evidence router.

\subsection{Router Structure}

The evidence router is a lightweight learned selector over serialized
structured scene--event units. Given a question \(q\) and a candidate memory unit
\(m_i\), the router predicts two node scores:
\(p_{\mathrm{sup}}(m_i\mid q)\), which estimates whether \(m_i\) contains
answer-supporting evidence, and \(p_{\mathrm{anc}}(m_i\mid q)\), which estimates
whether \(m_i\) grounds the event, state, source, or temporal transition
expressed by the question. For a pair of candidate units \((m_i,m_j)\), the
router predicts \(p_{\mathrm{edge}}(m_i,m_j\mid q)\), which estimates whether
the two units should be selected together as a directed anchor--support link.

The router uses hashed sparse features and three binary linear classifiers: a
support head, an anchor head, and an edge head. In our implementation, the
node-level feature space has dimension 2048, and the support, anchor, and edge
heads contain 1711, 1711, and 784 nonzero weights, respectively. We also include
relation features between candidates in the pairwise edge scorer, using a
1024-dimensional hashed relation-feature space. This design keeps routing
separate from answer generation: the router selects evidence before the reader
answers, while the fixed reader consumes only the rendered evidence pack.

\subsection{Router Features}

For each candidate unit, the pointwise feature map combines query tokens,
serialized unit tokens, token overlap, retrieval score, token cost, reciprocal
retrieval rank, step or turn position, exact step mentions, action or location
fields, objects, events, relations, and state keys when available.

For each candidate pair, the pairwise feature map includes temporal distance,
adjacency, order direction, shared query tokens, shared text, same-location
indicators, shared objects, same event type, and shared state keys. The relation
features further include temporal adjacency, temporal-near,
temporal-forward/backward direction, same-location, shared-object,
same-event-type, shared-state-key, and retrieval co-occurrence.




\subsection{Candidate Generation and Labels}

For each question, candidate memory units are retrieved with a hybrid lexical
and dense retrieval stage over serialized units. The candidate set contains the
top retrieved units. During supervised router-data construction, weakly labeled
units can be inserted into the candidate set when necessary so that training
labels are not lost due to retrieval misses. This insertion is used only for
constructing supervised router training data; it is not used as gold evidence
at benchmark inference time.

Each candidate stores its step identifier, serialized text, retrieval features,
token cost, and two binary labels:
\[
y^{\mathrm{sup}}_i=\mathbb{1}[t_i\in G_{\mathrm{sup}}], \qquad
y^{\mathrm{anc}}_i=\mathbb{1}[t_i\in G_{\mathrm{anc}}].
\]
The support and anchor heads are trained from all candidate nodes. The edge head
uses weak chain edges as positive pairs. Negative edge pairs are sampled from
adjacent candidate pairs, pairs involving support candidates, and random
non-positive candidate pairs, with at most 64 negatives per example. Positive
class weights are used for all heads to compensate for sparse support, anchor,
and edge labels.

\subsection{Router Training Example}

Table~\ref{tab:router-data-case} shows one router-training example. The question
requires counting two occurrences of the same action within a step range. Weak
support and anchor labels are assigned to the two matching memory units, while
high-overlap distractors become hard negatives.

\begin{table*}[t]
\centering
\small

\begin{tabular}{cllccr}
\toprule
Rank & Step & Action & \(y^{\mathrm{sup}}\) & \(y^{\mathrm{anc}}\) & Cost \\
\midrule
1 & 3 & \texttt{mix picture} & 1 & 1 & 33 \\
2 & 4 & \texttt{mix picture} & 1 & 1 & 33 \\
3 & 1 & \texttt{look at orange} & 0 & 0 & 29 \\
4 & 5 & \texttt{mix door to kitchen} & 0 & 0 & 37 \\
6 & 39 & \texttt{focus on hallway} & 0 & 0 & 109 \\
\bottomrule
\end{tabular}
\caption{Example router training instance. The question requires counting two
occurrences of the same action within a step range. Weak support and anchor
labels are assigned to the two matching memory units, while high-overlap
distractors become hard negatives.}
\label{tab:router-data-case}
\end{table*}

For the question ``From steps 1 to 39, how many times did you execute
\texttt{mix picture}?'' the answer is 2. The weak supervision marks steps 3 and
4 as support steps, because both match the queried action within the specified
range. They also serve as anchors because the question asks for an event count.
The directed edge label is therefore \((3,4)\), indicating that the two
occurrences form the evidence chain needed for counting.

The serialized positive units are:
\begin{unitlisting}{Serialized positive units}
[STEP] 3; [ACTION] mix picture; [LOCATION] hallway;
[OBS] That container is empty, so there are no items to mix;
[OBJECTS] picture, orange; [EVENTS] act action=mix picture.

[STEP] 4; [ACTION] mix picture; [LOCATION] hallway;
[OBS] That container is empty, so there are no items to mix;
[OBJECTS] picture, orange; [EVENTS] act action=mix picture.
\end{unitlisting}

The same retrieved candidate set also contains hard negatives, such as

\begin{unitlisting}{Serialized positive units}
\[
\texttt{[STEP] 5;\ [ACTION] mix door to kitchen;\ [LOCATION] hallway}.
\]
\end{unitlisting}

This candidate shares the token \texttt{mix}, the same location, and the same
visible objects, but it does not match the queried action
\texttt{mix picture}. This example illustrates why the router is trained with
support labels, anchor labels, and pairwise edge labels rather than relying only
on retrieval rank: lexical retrieval can produce plausible distractors, while
the structured labels identify the actual event chain.

\subsection{Training-Data Audit}

We audit the router-training data to separate weakly labeled examples from
unlabeled converted examples. The supervised router training set contains
\(2{,}868\) joint examples after candidate construction. Among the converted
environment-specific files, \(1{,}209\) examples in the Crafter-derived split,
\(242\) examples in the ScienceWorld-derived split, and \(1{,}086\) examples in
the EMemBench-derived file contain weak evidence labels. Converted AMA-Bench
examples are not used as supervised weak-label training instances in this audit.

\section{Weak Supervision and Router Training}
\label{app:weak_supervision}

\noindent\textbf{Training data sources.}
We construct router-training data from four interactive environments:
\crafter{}, \jericho{}, \sciworld{}, and \alfworld{}.
These environments provide heterogeneous long-horizon histories, including
visual game trajectories, text-game trajectories, science task trajectories,
and embodied instruction traces.
For each environment, we process completed episodes into trajectory records and
use GPT-5-assisted generation to standardize trajectory-grounded QA examples.
Each example contains a trajectory, a question, an answer, question metadata,
and weak evidence annotations when available. This construction is used only to
train the evidence router; it is not used to train the answer-time reader.

\begin{table*}[t]
\centering
\small

\begin{tabular}{lll}
\toprule
\textbf{Source} & \textbf{Raw history type} & \textbf{Router labels} \\
\midrule
\crafter{} & visual/game trajectories & support / anchor / edge \\
\jericho{} & text-game trajectories & support / anchor / edge \\
\sciworld{} & science task trajectories & support / anchor / edge \\
\alfworld{} & embodied instruction traces & support / anchor / edge \\
\bottomrule
\end{tabular}
\caption{
Weak-supervision sources for router training.
The constructed labels supervise evidence selection rather than answer generation.
}
\label{tab:weak_supervision_sources}
\end{table*}

\noindent\textbf{Label construction.}
Each trajectory step is first converted into a structured scene--event unit
using the schema in Sec.~\ref{sec:structured_units}. For each question, weak
evidence annotations are mapped to support labels over memory units. Question
metadata, such as answer type, target step, occurrence type, temporal relation,
or source constraint, is used to derive query-anchor labels. For questions whose
answers require multiple evidence steps, ordered evidence steps are converted
into anchor--support edge labels. The resulting data supervise three routing
targets: whether a unit is answer-supporting, whether a unit matches the query
anchor, and whether two units should be selected together in a short evidence
chain.

\noindent\textbf{Router architecture.}
The router is a lightweight scorer over structured unit features rather than an
answer generator. Given a question \(q\), a candidate unit \(m_i\), and a
candidate pair \((m_i,m_j)\), the router computes pointwise features
\(\phi(q,m_i)\) and pairwise features \(\psi(q,m_i,m_j)\). It then predicts
support, anchor, and edge scores:
\[
p_{\mathrm{sup}}(m_i\mid q)=\sigma(w_s^\top \phi(q,m_i)+b_s),
\]
\[
p_{\mathrm{anc}}(m_i\mid q)=\sigma(w_a^\top \phi(q,m_i)+b_a),
\]
\[
p_{\mathrm{edge}}(m_i,m_j\mid q)=\sigma(w_e^\top \psi(q,m_i,m_j)+b_e).
\]
These heads are trained offline from the weak labels above. At benchmark
inference time, all router parameters are fixed.

\noindent\textbf{Training configuration.}
We train the router only on the weakly supervised trajectory-QA data described
above. The writer, packer, and reader are frozen throughout training and
evaluation. We do not update the reader, use benchmark answers as supervision,
or train on target benchmark gold evidence.


\subsection{Reader-Model Sensitivity}
\label{app:reader_sensitivity}

We test whether the observed routing trend depends on a single answer-time
reader. The selected-evidence protocol and EMemBench text external subset are
held fixed, while the reader model is varied. Scores are reported in
Table~\ref{tab:reader_sensitivity_app}. Evidence-token counts are determined by
the memory method rather than by the reader.

\noindent\textbf{Analysis.}
\method{} is best under DeepSeek-v4-pro and Gemini3Flash, and remains close to
the strongest method under GLM5.1. HippoRAG is also strong, especially under
GLM5.1, but it uses a larger evidence context in this setting. These results
indicate that the routing trend is not tied to a single answer-time reader,
while also showing that reader choice can affect absolute scores.

\section{Evaluation Protocol and Leakage Control}
\label{app:protocol}
\label{app:leakage_control}

This section specifies the evaluation boundary used in the main paper.

\subsection{No-Gold Inference Boundary}

For the main benchmark results, inference follows a no-gold boundary.  The method may access the test question and the corresponding memory source, but it does not access the gold answer, gold evidence, judge labels, benchmark scores, or the reader output before evidence selection.  Evidence is selected before the reader answers.



We do not treat an adapted baseline as an exact reproduction of the full original system.  Adapted comparisons are used to compare reader-visible evidence quality under a shared evaluation interface.

\subsection{Leakage-Control Rules}

We follow four rules to reduce leakage risk:
\begin{enumerate}
    \item The writer receives raw history records only and is question-independent unless otherwise stated.
    \item The router receives the test question and written memory units, but not benchmark answers or judge feedback.
    \item The packer uses router scores and structured-unit fields only; it does not consult reader outputs.
    \item Diagnostic analyses that use additional information are reported separately from the main learned-router results and are not used to support the main no-gold inference claim.
\end{enumerate}

\section{Benchmark Conversion, Scoring, and Baseline Provenance}
\label{app:benchmark_conversion}
\label{app:api_tokens}

\subsection{Benchmark Details}

LoCoMo~\citep{locomo2024} is used for text memory evaluation. We follow the
question categories used in the main table, including temporal, multi-hop,
single-hop, open-domain, and adversarial questions. The weighted overall score
is computed according to the number of questions in each category.

EMemBench Visual Games~\citep{emembench2026} is used for visual memory
evaluation. The benchmark requires evidence from frame-level observations,
object states, event occurrences, and visual state changes.

AMA-Bench~\citep{amabench2026} is used for agent-use memory evaluation. The
benchmark includes Recall, Causal Inference, State Updating, and State
Abstraction categories over agent trajectories.




\subsection{Token Accounting}

We report tok/q as the average number of evidence tokens exposed to the
answer-time reader per question. This metric measures the compactness of the
memory interface shown to the reader. It does not include all preprocessing,
writing, routing, or API usage costs. Therefore, the score--token analysis in
the main paper should be interpreted as a reader-visible evidence-efficiency
comparison, not as a total serving-cost comparison.

\subsection{Benchmark Conversion}

For each benchmark, raw memory sources are converted into the common structured scene--event unit schema before routing.  Text-memory examples are converted from long conversational histories or textual records.  Visual-memory examples are converted from frame-level observations, object states, and event traces.  Agent-use examples are converted from action trajectories, observations, state updates, and task metadata.

Benchmark-specific code is limited to input conversion, prompt rendering, and official or benchmark-compatible scoring.  The router architecture is not changed across benchmarks.

\subsection{Scoring Protocols}

For LoCoMo, we report F1 and BLEU by question category and compute weighted overall scores using the category counts reported in the main table.  For EMemBench Visual Games, we report accuracy, F1, and average reader-visible evidence tokens per question.  For AMA-Bench, we report accuracy on Recall, Causal Inference, State Updating, and State Abstraction, together with their average.

\subsection{Reader-Visible Evidence Tokens}

We report \textbf{tok/q} as the average number of evidence tokens exposed to the answer-time reader per question.  This measures the size of the memory interface shown to the reader.  It should not be interpreted as total system API usage, because writing, candidate proposal, routing, and optional preprocessing can incur additional cost.  The main claim concerns the compactness and usefulness of reader-visible evidence, not total end-to-end serving cost.










\section{Supplemental Results and Diagnostic Analyses}
\label{app:supp_results}

This section contains supplemental results that clarify the main findings.  These analyses should be interpreted as diagnostics rather than as replacements for the main benchmark tables.

\subsection{Representation Ablation}

The representation ablation in the main paper compares the full structured scene--event representation with plain chunks and single-field variants.  The result supports a bounded conclusion: on EMemBench External, the full structured unit is more effective than using plain text chunks or isolated event/object fields.  This ablation does not by itself prove that every router component is necessary.

\subsection{Recommended Router Ablations}

To isolate the contribution of routing components, we recommend reporting the following ablations when available:
\begin{itemize}
    \item \textbf{Full \method{}}: structured units with support, anchor, and edge scores.
    \item \textbf{w/o query anchors}: candidate selection and packing without anchor scores.
    \item \textbf{w/o edge scores}: single-unit support scoring without anchor--support links.
    \item \textbf{Top-\(k\) support only}: selects top-scoring units without structured packing.
    \item \textbf{Plain chunks}: retrieves and renders unstructured text chunks.
\end{itemize}

If these rows are not included, claims about the necessity of query anchors, edge scoring, or packing should be stated as design motivation rather than experimentally established conclusions.

\subsection{Reader Sensitivity}

Reader-sensitivity experiments can help determine whether the routing trend depends on one answer-time model.  When reporting such results, the table should specify the reader, evidence-token accounting, and whether the same selected evidence was used across readers.  The interpretation should focus on robustness of the evidence interface, not on claiming a universal reader-independent improvement.

\section{Qualitative Case Studies}
\label{app:case_studies}

The following examples illustrate how compact structured evidence can help in cases where broad retrieval or compression may return incomplete or distracting evidence.  They are illustrative only and do not replace aggregate benchmark results.

\begin{casebox}{Example A: Visual Temporal Interval Reasoning}
\textbf{Question.}
After the step when your water first dropped below 3, after how many steps did the step when your health first became 2 or lower occur?

\medskip
\noindent
\textbf{Gold answer.} 15

\medskip
\noindent
\textbf{Selected evidence.}
The evidence pack keeps the first event where water drops below 3 and the later event where health becomes 2 or lower:
\[
\begin{aligned}
\text{step }147 &: \text{water}=2,\ \text{health}=4,\\
\text{step }162 &: \text{health}=1,\ \text{water}=2.
\end{aligned}
\]
The reader can then compute \(162-147=15\).

\medskip
\noindent
\textbf{Why this case matters.}
The question requires two threshold-crossing events and their temporal distance.  A related state summary is not sufficient unless it preserves both events and their step indices.
\end{casebox}

\begin{casebox}{Example B: Visual Counting of Repeated Events}
\textbf{Question.}
How many times were you assaulted?

\medskip
\noindent
\textbf{Gold answer.} 6

\medskip
\noindent
\textbf{Selected evidence.}
The evidence pack keeps repeated attack events as separate units rather than merging them into one summary.  The support units include repeated attack events with associated health changes.

\medskip
\noindent
\textbf{Why this case matters.}
Counting questions can fail when compression collapses repeated low-level events as redundant.  Structured evidence packing can preserve repeated occurrences as separate countable units.
\end{casebox}

\begin{casebox}{Example C: Before/After Step Boundary}
\textbf{Question.}
At step 4, what observation did you see before performing the action?

\medskip
\noindent
\textbf{Gold answer.}
You switch the brass lantern on.

\medskip
\noindent
\textbf{Selected evidence.}
The evidence pack includes the neighboring step before the queried action: step 3 contains the action \texttt{turn lantern on} and the observation \texttt{You switch the brass lantern on.}

\medskip
\noindent
\textbf{Why this case matters.}
The decisive evidence is not merely the presence of related words such as \texttt{lantern}.  The reader needs the observation at the correct temporal boundary.
\end{casebox}

\section{Failure Cases and Claim Boundary}
\label{app:limitations}

\subsection{Failure Modes}

\method{} can fail in several ways.  First, the router may under-select evidence, producing a compact but incomplete support chain.  Second, it may identify the correct anchor but miss a required support step.  Third, the reader may fail even when the selected evidence contains the needed information.  Fourth, some benchmark questions may require broad recall or high-level abstraction over long trajectories, where compact routing is less effective than broader graph traversal or larger-context methods.

\subsection{Supported Claims}

The experiments support the following bounded claims:
\begin{itemize}
    \item Structured scene--event units can provide useful reader-facing evidence for long-horizon memory QA.
    \item Routing over support, anchor, and edge signals provides a compact evidence interface for fixed readers.
    \item \method{} improves the score--token trade-off in the evaluated settings, with the clearest gains on localized event, state, temporal, causal, and provenance evidence.
\end{itemize}



\section{Structured Memory Writer Prompt}
\label{app:writer_prompt}

We use a frozen LLM as the structured memory writer.  The writer receives raw history records and emits structured scene--event units in JSON format.  It does not receive benchmark questions, gold answers, gold evidence, judge labels, or benchmark scores.

\begin{tcolorbox}[title=Structured Memory Writer Prompt]
You are a memory writer for a long-horizon question-answering system. Convert the raw history record into structured scene--event memory units.

For each unit, output a JSON object with the following fields:
\begin{itemize}
    \item \texttt{time}: timestamp, step id, frame id, or turn id.
    \item \texttt{event}: concise description of the event or observation.
    \item \texttt{objects}: entities, objects, tools, people, or variables involved.
    \item \texttt{location\_source}: location, document, frame, conversation, tool trace, or source context.
    \item \texttt{state\_delta}: state before/after the event, if available.
    \item \texttt{relations}: temporal, causal, object-state, visual, or provenance links to other units.
    \item \texttt{provenance}: source metadata needed to verify the unit.
    \item \texttt{raw\_pointer}: pointer to the original record.
\end{itemize}

Rules:
\begin{itemize}
    \item Preserve factual details from the raw record.
    \item Do not infer unsupported facts.
    \item Keep each unit concise.
    \item If a field is unavailable, use \texttt{null} or an empty list.
    \item Return only valid JSON.
\end{itemize}
\end{tcolorbox}


\end{document}